\documentclass[conference]{IEEEtran}

\usepackage{pgfplots}
\usepackage{graphicx}
\usepackage{color}
\usepackage{tikz}
\usepackage{pgfplots}
\usepackage{pgf-umlsd}
\usepackage{ifthen}
\usepackage{cite}

\newcommand{\LABTAB}[1]{\label{tab:#1}}
\newcommand{\LABFIG}[1]{\label{fig:#1}}
\newcommand{\TAB}[1]{{Table~\ref{tab:#1}}}

\newcommand{\FIG}[1]{Fig.~\ref{fig:#1}} 

\begin{document}

\title{Boosting Handwriting Text Recognition\\ in Small Databases with Transfer Learning}

\author{\IEEEauthorblockN{Jos\'e Carlos Aradillas}
\IEEEauthorblockA{University of Seville\\
Seville, Spain 41092\\
Email: jaradillas@us.es}
\and
\IEEEauthorblockN{Juan Jos\'e Murillo-Fuentes}
\IEEEauthorblockA{University of Seville\\
Seville, Spain 41092\\
Email: murillo@us.es}
\and
\IEEEauthorblockN{Pablo M. Olmos}
\IEEEauthorblockA{University Carlos III in Madrid\\
Madrid, Spain\\
Email: olmos@tsc.uc3m.es}}



%


\maketitle

\begin{abstract}
In this paper we deal with the offline handwriting text recognition (HTR) problem with reduced training data sets. Recent HTR solutions based on artificial neural networks exhibit remarkable solutions in referenced databases. These deep learning neural networks are composed of both convolutional (CNN) and long short-term memory recurrent units (LSTM). In addition, connectionist temporal classification (CTC) is the key to avoid segmentation at character level, greatly facilitating the labeling task. One of the main drawbacks of the CNN-LSTM-CTC (CLC) solutions is that they need a considerable part of the text to be transcribed for every type of calligraphy, typically in the order of a few thousands of lines. Furthermore, in some scenarios the text to transcribe is not that long, e.g. in the Washington database. The CLC typically overfits for this reduced number of training samples. Our proposal is based on the transfer learning (TL) from the parameters learned with a bigger database. We first investigate, for a reduced and fixed number of training samples, 350 lines, how the learning from a large database, the IAM, can be transferred to the learning of the CLC of a reduced database, Washington. We focus on which layers of the network could be not re-trained. We conclude that the best solution is to re-train the whole CLC parameters initialized to the values obtained after the training of the CLC from the larger database. We also investigate results when the training size is further reduced. For the sake of comparison, we study the character error rate (CER) with no dictionary or any language modeling technique. The differences in the CER are more remarkable when training with just 350 lines, a CER of 3.3\% is achieved with TL while we have a CER of 18.2\% when training from scratch. As a byproduct, the learning times are quite reduced. Similar good results are obtained from the Parzival database when trained with this reduced number of lines and this new approach.
\end{abstract}

\IEEEpeerreviewmaketitle

\section{Introduction}
Handwriting text recognition (HTR) on historical documents is a well-known necessary task for the preservation of the cultural heritage. Over the last decade,  there has been a trend towards the utilization of recurrent neural networks (RNNs) \cite{Graves09, Pham14, Voigtlaender16} joint with connectionist temporal classification (CTC) \cite{Graves06} in order to have an end-to-end system capable of doing the transcription of raw images containing whole lines of text. RNN-CTC methods for HTR have obtained the lowest error rates in recent HTR contests \cite{Graves09, Pham14, Voigtlaender16,  Bluche16a, Bluche16b, Grosicki09}. At the core of these methods,  we find  multidimensional  long short-term memory (MDLSTM) networks, which generalize standard RNNs by providing recurrent connections along each dimension in the data \cite{Graves09}. The resulting structure, while robust to local distortions along any combination of input dimensions, is hard to train due to the vast number of parameters and memory requirements to perform back-propagation through recurrent connections \cite{Voigtlaender16}.

Recently, a deep neural network for HTR including convolutional neural networks (CNNs) as a first stage has been presented in \cite{Puigcerver17}. Spatial correlation is tackled by the CNN, and bi-directional LSTMs \cite{Hoch96} (BLSTM) at the top of the network induce temporal correlations. No MDLSTM layers are then needed, since CNNs are able to extract latent representations of images that are robust to local distortions. Consequently, training is more efficient in terms of the number of parameters and memory requirements. The overall HTR structure presented in \cite{Puigcerver17} is summarized in \FIG{structure}. 

In this paper, we put forward this approach for HTR to apply transfer learning (TL) across different databases. With a smaller number of parameters and good generalization performance, we show that the network in \FIG{structure} is able to achieve remarkable generalization results for small historical databases once it has been pre-trained over a large database. We use the IAM database  \cite{Marti02} to pre-train the network, then use this learning to solve HTR in the Parzival  database \cite{Fisher12} by using only 350 lines of text. We show that a test character error rate (CER) as low as 3.3\% can be achieved. Similar results are also achieved for the Washington database \cite{Fisher12}. Furthermore, both databases are binarized, hence making the HTR more challenging. Up to our knowledge, transfer learning is a novel approach to HTR of text lines. In \cite{Granet18} they use TL for HTR to face a different problem in which the objective dataset has no ground-truth.

Results demonstrate the importance of performing TL as the right way to train HTR solutions based on deep neural networks that are able to generalize well over small databases. This is of critical importance in historical documents, typically characterized by small databases and a huge variety of calligraphic styles.

\begin{figure*}[!t]
\centering
\includegraphics[width=6.5in]{{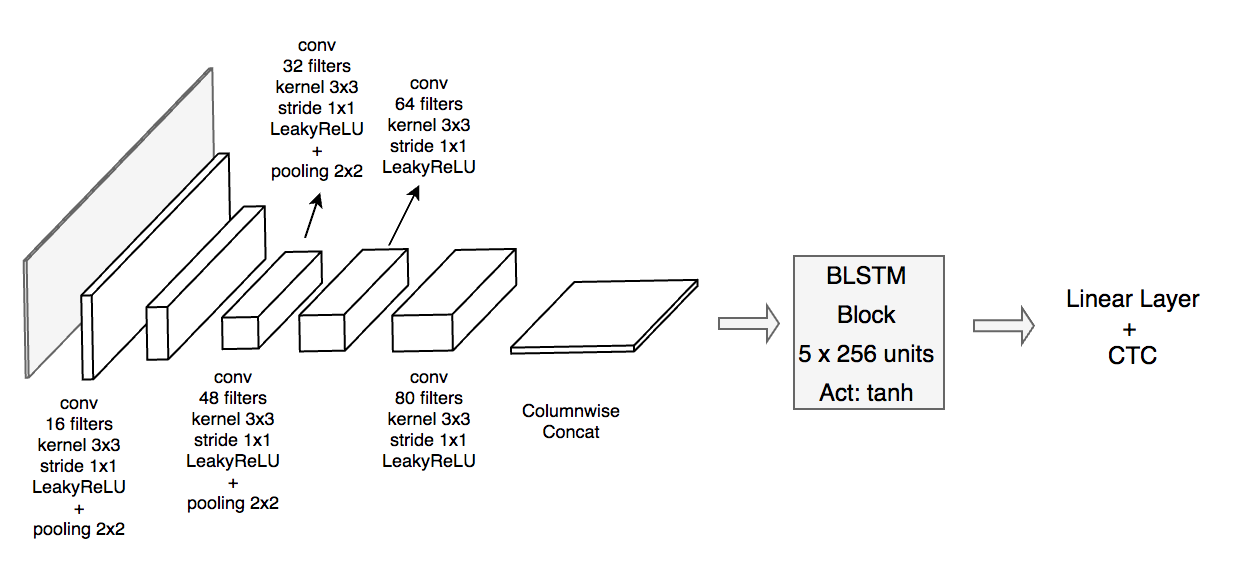}}%
\caption{The network architecture used in this paper.}
\LABFIG{structure}
\end{figure*}

The rest of the paper is organized as follows: Section \ref{sec:TL} provides an introduction to transfer learning; the databases used in this paper are presented in Section \ref{sec:databases};  the neural network trained in this paper is detailed in Section \ref{sec:architecture}; in Section \ref{sec:method} we analyze the application of transfer learning to solve HTR tasks over small databases and finally, conclusions are discussed in Section \ref{sec:conclusions}.

 

\section{Transfer Learning}\label{sec:TL}
In HTR tasks, deep learning algorithms have been usually focused on solving a problem over a domain $\mathcal{D} = \{\mathcal{X}, P(X)\}$ consisting of segmented images. The task consists of two components: a label space $\mathcal{Y}$  and an objective predictive function $f(\cdotp)$ (denoted by $\mathcal{T} = \{\mathcal{Y},f(\cdotp)\}$), which can be learned from the training data, which consists of pairs $\{x_i, y_i\}$, where $x_i \in \mathcal{X}$ and $y_i \in \mathcal{Y}$ \cite{Pan10}.
\\
Given a source domain $\mathcal{D}_S$ and a learning task $\mathcal{T}_S$, transfer learning aims to help improve the learning of another target predictive function $f_T(\cdotp)$ in $\mathcal{D}_T$ using the knowledge in $\mathcal{D}_S$ and  $\mathcal{T}_S$. In this work we are interested in \emph{inductive transfer learning} in which the target task is different from the source task, as the domains are different ($\mathcal{D}_S\neq \mathcal{D}_T$). In this work the source domain will be the IAM database \cite{Marti02}, while the target domain will be the Washington and Parzival databases \cite{Fisher12}.

\section{Databases}\label{sec:databases}
In this paper we focus on the HTR of three databases: IAM \cite{Marti02}, Washington and Parzival \cite{Fisher12}.

\subsection{The IAM database}
The IAM database \cite{Marti02} contains $13353$ labeled text lines of modern English handwritten by $657$ different writers. The images were scanned at a resolution of $300$ dpi and saved as PNG images with $256$ gray levels. An image of this database is included in \FIG{IAMsample}. The database is partitioned into training, validation and test sets of $6161$, $900$ and $2801$ lines, respectively\footnote{ The names of of the images of each set are provided in the \emph{Large Writer Independent Text Line Recognition Task}.}. Here, the validation set and test set provided are merged in a unique test set.
There are $79$ different characters in this database, including capital and small letters, numbers, some punctuation symbols and the white-space. 
\begin{figure}[!t]
\centering
\includegraphics[width=2.5in]{{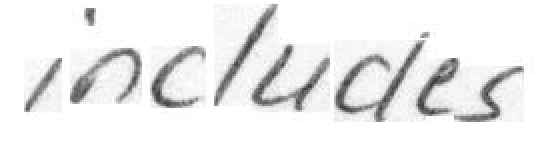}}%
\caption{IAM handwritten text sample}
\LABFIG{IAMsample}
\end{figure}

\subsection{The Washington database}
The Washington database contains $565$ text lines from the George Washington papers, handwritten by two writers in the 18th century. Although the language is also English, the text is written in longhand script and the images are binarized as illustrated in \FIG{Washingtonsample}, see \cite{Yousefi15} for a description of the differences between binarized and binarization-free images when applying HTR tasks. In this database four possible partitions of the whole are provided in order to train and validate. In this work we have chosen one of them randomly in order to train and validate the system. The train, validation and test set contain $325$, $168$ and $163$ handwritten lines, respectively. There are $83$ different characters in the database.

\begin{figure}[!t]
\centering
\includegraphics[width=2.5in]{{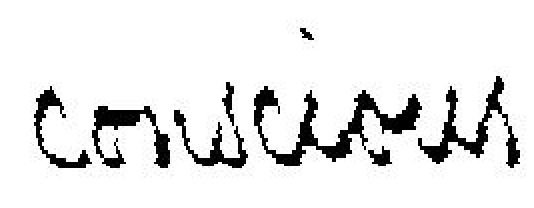}}%
\caption{Washington handwritten text sample}
\LABFIG{Washingtonsample}
\end{figure}

\subsection{The Parzival database}
The Parzival database contains $4477$ text lines handwritten by three writers in the 13-th century. In this case, the lines are binarized like in the Washington database, but the text is written in gothic script. A sample is included in \FIG{Parzivalsample}. There are $96$ different characters in this database. Note that the Parzival database has a large number of text lines in comparison to the Washington one.  We randomly choose a training set of the approximately same size that in the Washington training to emulate learning with a small dataset, the main goal of this work. 

\begin{figure}[!t]
\centering
\includegraphics[width=2.5in]{{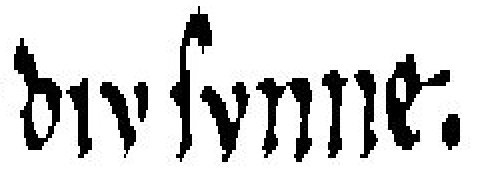}}%
\caption{Parzival handwritten text sample}
\LABFIG{Parzivalsample}
\end{figure}
\section{Architecture}\label{sec:architecture}

In this work, it is proposed to use the architecture presented in \cite{Puigcerver17} and shown in \FIG{structure} which avoids the use of MDLSTM layers and applies convolutional layers as feature extractors and a stack of 1D BLSTM layers to perform the classification. Previous deep neural network architectures for HTR proposed architectures consisting of a combination of MDLSTM layers and convolutional layers, with a collapse phase before the output layer in order to reshape the features matrix from 2D to 1D \cite{Voigtlaender16, Pham14}. The use of 2D-LSTM layers at the first stages has several drawbacks since it requires more memory for the allocation of activations and buffers during back-propagation, and the runtime required to train the networks is higher. Recently, it has been proven that CNN in the lower layers of an HTR system obtains similar features than a RNN containing 2D-LSTM units \cite{Puigcerver17}.

Note that the higher layers in the structure shown in \FIG{structure} implement BLSTM. The neural network is composed by $5$ convolutional layers CNNs all with a kernel of $3\times3$ and $1\times1$ stride, the number of filters are $16$, $32$, $48$, $64$ and $80$, respectively. We use LeakyReLU as activation function. A $2\times2$ max-pooling is also applied at the output of the first $3$, with the aim of reducing the size of the input sequence. After this first convolutional stage, a column-wise concatenation is carried out with the purpose of transforming the 3D tensors of size (width $\times$ height $\times$ depth) into 2D tensors of size (width $\times$ (height $\times$ depth)), therefore at the entrance of the first layer of BLSTMs we have sequences of length equal to the width of the image after having been applied 3 stages of $2\times2$ max-pooling and a number of features equal to $80\times H$, where $H$ is the height of the images after pooling stages.

After the CNN stage, $5$ BLSTM recurrent layers of $256$ units without peepholes connections and hyperbolic tangent activation functions are applied. Since at the output of each BLSTM layer we have $256$ features for each direction, we perform a depth-wise concatenation in order to adapt the input of the next layer, in overall size of $512$. Dropout regularization \cite{Pham14, Srivastava14} is applied at the output of every layer except for the first convolutional one with rates $0.2$ for the CNN layers and $0.5$ for the LSTM layers.

Finally, each column of features after the 5th BLSTM layer, with depth $512$, is mapped to an output label with a full connected (FC) network. $L + 1$ labels are used, where $L$ is the number of characters that appears in the ground truth of each database, $79$ for IAM database, $83$ in the Washington database and $96$ in the Parzival database. The additional dimension is needed for the blank symbol of the CTC \cite{Graves06}, that ends this architecture. Overall, this CNN-BLSTM-CTC architecture has a total number of to-be-learned parameters of $9581008$ for IAM, $9589729$ for Parzival and $9583060$ for Washington,  where the differences are due to the number of characters in each database.

\subsection{Implementation}
The architecture is implemented in the open source framework TensorFlow in Python, over their GPU version. We use the Adam algorithm \cite{Kingma14}, a learning rate of $0.003$. The parameters are updated using the gradients of the CTC loss on each batch of $20$ text lines\footnote{The code will be available in https://github.com/josarajar/HTRTF }.

\section{Transfer Learning for HTR}\label{sec:method}

In this paper we investigate how TL can be applied to reuse the parameters learned during the training of a large database for the learning of another reduced corpus in the HTR problem. While the large database has thousands of lines ($5000 - 7000$) the training set for the new HTR problem is of size a few hundreds ($150-350$). 
The proposed methodology is as follows: We first studied in detail what the best strategy to perform TL might be. We did TL from the IAM database to the Washington one, first with $325$ text lines as training data to later face the learning with $250$ and $150$ text lines. Then, the best strategies found were validated by the Parzival database. 

\subsection{Learning from scratch }
 When training the CNN-BLSTM-CTC architecture from scratch for the Washington database, the CER tends to 0\% if it is evaluated over the training set while a CER of over 40\%  is reached when evaluating the validation set. It can be concluded that there is not enough number of samples and  we have overfitting. The convergence is depicted in \FIG{CERcurve}.
\\
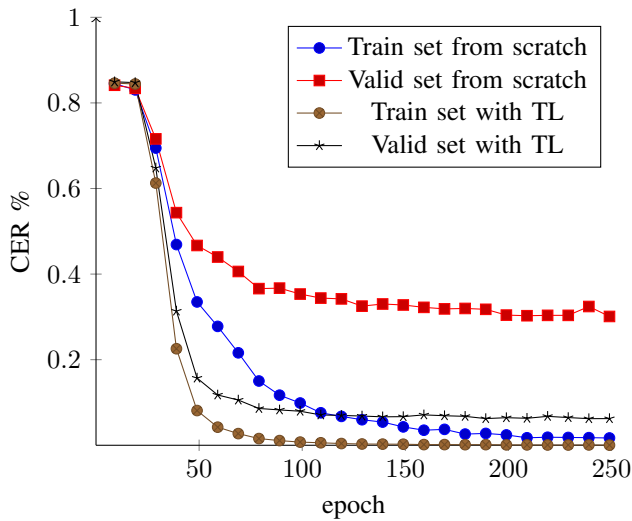
\begin{figure}[!t]
\centering

\begin{tikzpicture}
\begin{axis}[ 
    axis lines=middle,
    axis line style={->},
    ylabel near ticks,
    xlabel near ticks,
    ymin=0, ymax=1, xmin=0, xmax=250,
    xlabel={epoch},
    ylabel={CER \%}]
\addplot table [x=Step, y=Value, col sep=comma] {scratchtrain.csv};
\addplot table [x=Step, y=Value, col sep=comma] {scratchvalid.csv};
\addplot table [x=Step, y=Value, col sep=comma] {TLtrain.csv};
\addplot table [x=Step, y=Value, col sep=comma] {TLvalid.csv};
\addlegendentry{Train set from scratch}
\addlegendentry{Valid set from scratch}
\addlegendentry{Train set with TL}
\addlegendentry{Valid set with TL}
\end{axis}
\end{tikzpicture}
\caption{Evolution of the error while training the CNN-BLSTM-CTC with random initialization and the Washington database.
}
\LABFIG{CERcurve}
\end{figure}

\subsection{Simple TL by just initialization}
In the first instance, we trained the network for a large database, IAM, and applied the learned solution to the Washington database. We only transcribed the common characters to both databases. We got a CER = 82\%. This poor result may be due to the heterogeneity between both databases: 1) their images have different resolution, 2) the calligraphies of the texts correspond to different centuries, 3) the sets or alphabets of characters are different and 4) the images used in the training are in gray scale while the database where the learning results are applied is binarized. These differences can be observed in \FIG{IAMsample} and \FIG{Washingtonsample}. At this point, it is interesting to note that due to the stage of column-wise concatenation between CNN and BLSTM layers, the number of parameters in the first BLSTM layer depends on the height of the input images. Therefore, in order to apply the same CNN-BLSTM-CTC structure to two different databases, it is necessary to resize the images of the target database so that they have the same height as the images of the training database.

\subsection{Best TL strategy}
As a first approach of TL we added one more BLSTM layer on top of the BLSTM part of the CNN-BLSTM-CTC architecture. We learned this new layer, while keeping the rest of layers fixed to the values learned for the IAM database. This architecture is prone to overfitting. However, reducing the number of units of this new layer from $256$ to $128$ or $64$ did underfit. These results discouraged us from adding new layers to the already trained architecture.

\begin{table}[!t]
\renewcommand{\arraystretch}{1.3}
\caption{Character error rate (CER \%) evaluated in Washington datasets in a model obtained after retraining a set of layers in the model previously trained over the IAM database. It has been retrained with 325 lines images}
\LABTAB{Washington325}
\centering
\begin{tabular}{lrrr}
\hline
& \multicolumn{3}{c}{CER (\%)} \\
Trainable Layers & Train & Valid & Test\\
\hline
FC & 55.1 & 46.2 & 47.1  \\
BLSTM5, FC & 13.0 & 22.1 & 23.6    \\
BLSTM[4,5], FC & 4.2 & 14.4 & 17.4  \\
BLSTM[3,4,5], FC & 0.6 & 10.6 & 12.8   \\
BLSTM[2,3,4,5], FC & 0.2 & 8.7 & 6.7   \\
BLSTM[1,2,3,4,5], FC & 0.2 & 5.3 & 6.6    \\
Conv[5], BLSTM[1,2,3,4,5], FC & 0.2 & 4.8 & 6.1    \\
Conv[4,5]. BLSTM[1,2,3,4,5], FC & 0.2 & 5.2 & 6.3    \\
Conv[3,4,5], BLSTM[1,2,3,4,5], FC & 0.3 & 4.5 & 5.5    \\
Conv[2,3,4,5], BLSTM[1,2,3,4,5], FC & 0.5 & \textbf{4.3} & 5.4    \\
Conv[1,2,3,4,5], BLSTM[1,2,3,4,5], FC & 0.2 & 4.6 & \textbf{5.3}    \\
\hline
\end{tabular}
\end{table}
We next kept the architecture and studied how the parameters could be initialized and learned. We initialized the architecture to the values trained with the IAM database, and then trained just a subset of layers from top to bottom, using $325$ text lines from the Washington database. The result of this analysis can be observed in \TAB{Washington325}. This table includes the CER evaluated on the training, evaluation and test sets. In the first column of the table, we indicate the layers that have been left free during the re-training, the rest of the layers remain fixed and initialized to the result of the learning of the large database. For example, BLSTM[3,4,5] FC indicates that the three upper BLSTM and the FC layers have been retrained, keeping the rest of the network fixed. The analysis involves both the layers in the BLSTM networks and the CNN layers.

We first trained just the FC layer of the structure in order to include the number of characters that the Washington database has ($83 + 1$ for CTC blank character). We re-trained this last layer with a training set of 325 lines from the Washington database, keeping the rest of the layers fixed. In \TAB{Washington325}, first row, it can be observed that just retraining the FC layer, the model tends to underfit, a CER = 52\% is obtained, evaluated both in the train set and in the validation set. We next included  the BLSTMs in the set of layers to be re-trained. See rows 2-6 in  \TAB{Washington325}.  Then we also re-trained the CNNs, rows 7 to 11.

The most interesting conclusion of this analysis is that by retraining the first four BLSTM network the CER decreases from $47$ \% to $6.7$ \%. By re-training the convolutional layers, i.e. the whole network,  we get an extra gain from $6.7$ \% to $5.3$ \%. The best result in the test set was obtained when the whole model is retrained (CER = $5.3$ \%) while in the validation set was obtained when the first convolutional layer is kept fixed (CER = $4.3$ \%). Also, by just re-training the top three convolutional layers we already get $5.5$ \%. A possible interpretation of this behavior is as follows: While CNNs are extracting features from the images \cite{Bluche13}, the BLSTM are supporting the classification task. The feature extraction stage is more transferable, the classification step is not.   

From this analysis we can conclude that a good TL approach would be to initialize and re-train the whole network. However, if the first two CNN layers are kept fixed we get approximately the same result.

\subsection{Reducing the training set }
We also investigate the performance of this proposed TL approach when the set of training data was reduced from $325$ to $250$ and $150$. 
We first randomly chose 250 text lines and repeated the same analysis that it was done over the set of $325$ lines to investigate the number of layers that should be retrained. The results of this analysis are included in \TAB{Washington250}. In this case, the best error rate was achieved when the whole model except the first convolutional layer was re-trained for both validation and test sets, (CER = $6.0$ \% and CER = $7.1$ \% respectively). The results are slightly worse than those of the model trained with $325$ text lines. To illustrate the convergence of the proposed TL, we include the convergence along epochs for this case in \FIG{CERcurve}, re-training the whole network.

Finally, we reduced the training set to $150$ text lines and repeated the experiments. The results of this analysis are included in \TAB{Washington150}. In this case, the best result for the validation set was achieved when the whole model was retrained fixing the two lower convolutional layers (CER = $7.9$ \%) while in the test set was achieved when keeping just the first layer fixed (CER = $9.4$ \%). 
These results could be considered promising if the cost of manually annotating the lines when creating the training dataset is taken into account \cite{Serrano10} \footnote{The total time required for a single expert to manually annotate 20357 lines was estimated as 500 hours}. It can also be concluded that not re-training the first, or the first and second, CNN layer is a robust strategy with varying training data size.
\begin{table}[!t]
\renewcommand{\arraystretch}{1.3}
\caption{Character error rate (CER \%) evaluated in Washington datasets in a model obtained after retraining a set of layers in the model previously trained over the IAM database. It has been retrained with $250$ lines images}
%

\LABTAB{Washington250}
\centering
\begin{tabular}{lrr}
\hline
& \multicolumn{2}{c}{CER (\%)} \\
Trainable Layers &  Validation & Test\\
\hline
BLSTM5, FC &  26.0 & 26.9    \\
BLSTM[4,5], FC &  18.9 & 19.5  \\
BLSTM[3,4,5], FC &  12.2 & 14.4   \\
BLSTM[2,3,4,5], FC &  8.4 & 10.5   \\
BLSTM[1,2,3,4,5], FC &  7.1 & 8.4    \\
Conv[5], BLSTM[1,2,3,4,5], FC &  6.6 & 8.1    \\
Conv[4,5]. BLSTM[1,2,3,4,5], FC &  5.8 & 7.3    \\
Conv[3,4,5], BLSTM[1,2,3,4,5], FC &  6.2 & 7.2    \\
Conv[2,3,4,5], BLSTM[1,2,3,4,5], FC &  \textbf{6.0} & \textbf{7.1}    \\
Conv[1,2,3,4,5], BLSTM[1,2,3,4,5], FC &  6.2 & 7.6    \\
\hline
\end{tabular}
\end{table}

\begin{table}[!t]
\renewcommand{\arraystretch}{1.3}
\caption{Character error rate (CER \%) evaluated in Washington datasets in a model obtained after retraining a set of layers in the model previously trained over the IAM database. It has been retrained with 150 lines images}
\LABTAB{Washington150}
\centering
\begin{tabular}{lrr}
\hline
& \multicolumn{2}{c}{CER (\%)} \\
Trainable Layers &  Validation & Test\\
\hline
BLSTM5, FC &  30.5 & 31.4    \\
BLSTM[4,5], FC &  22.7 & 24.2  \\
BLSTM[3,4,5], FC &  15.7 & 18.6   \\
BLSTM[2,3,4,5], FC &  11.4 & 14   \\
BLSTM[1,2,3,4,5], FC &  10.3 & 12.6    \\
Conv[5], BLSTM[1,2,3,4,5], FC &  9.2 & 11.2    \\
Conv[4,5]. BLSTM[1,2,3,4,5], FC &  8.1 & 10.1    \\
Conv[3,4,5], BLSTM[1,2,3,4,5], FC &  \textbf{7.9} & 9.5    \\
Conv[2,3,4,5], BLSTM[1,2,3,4,5], FC &  8.4 & \textbf{9.4}    \\
Conv[1,2,3,4,5], BLSTM[1,2,3,4,5], FC &  10.4 & 11.9    \\
\hline
\end{tabular}
\end{table}

\subsection{Validation with the Parzival database}
To validate the results obtained on the proposed TL algorithm, we applied it to the Parzival database. The Parzival database contains more than $2000$ annotated text lines. We randomly chose a reduced subset of $350$, $250$ and $150$ text lines to perform the TL. The CNN-BLSTM-CTC architecture trained from scratch with this 2000 lines train set achieves a CER = $1.7$ \% for both validation and test sets. Same model trained for $350$ lines got stuck in a value of CER = $18.2$ \%, similar to the value of this architecture for the Washington database. 

In the view of the previous results, we focused on TL re-training just the CNN layers. The results can be observed in \TAB{Parzival350}. In this case the best results were also obtained when the whole model or just the first layer was kept fixed: CER = $3.0$ \% and $3.3$ \% in the validation and test set, respectively, in both cases. 

When reducing the number of lines similar results were obtained. We applied TL by training a model with a training set of 250 and 150 images respectively. The results of these analyses are included in \TAB{Parzival250} and \TAB{Parzival150}. As in the Washington case, the best error rates are obtained when retraining the whole model or at most keeping fixed the lower CNN layers. The best results in this case are CER = $4.0$ \% for the training with $250$ text lines and CER = $5.8$ \% for the $150$ text lines case.

\begin{table}[!t]
\renewcommand{\arraystretch}{1.3}
\caption{Character error rate (CER \%) evaluated in Parzival datasets in a model obtained after retraining a set of layers in the model previously trained over the IAM database. It has been retrained with a set of 350 lines images}
\LABTAB{Parzival350}
\centering
\begin{tabular}{lrr}
\hline
& \multicolumn{2}{c}{CER (\%)} \\
Trainable Layers &  Validation & Test\\
\hline
BLSTM[1,2,3,4,5], FC &  4.2 & 4.1   \\
Conv[5], BLSTM[1,2,3,4,5], FC &  3.8 & 3.8    \\
Conv[4,5]. BLSTM[1,2,3,4,5], FC &  3.3 & 3.6     \\
Conv[3,4,5], BLSTM[1,2,3,4,5], FC &  3.4 & 3.5     \\
Conv[2,3,4,5], BLSTM[1,2,3,4,5], FC &  \textbf{3.0} & \textbf{3.3}   \\
Conv[1,2,3,4,5], BLSTM[1,2,3,4,5], FC &  \textbf{3.0} & \textbf{3.3}    \\
\hline
\end{tabular}
\end{table}

\begin{table}[!t]
\renewcommand{\arraystretch}{1.3}
\caption{Character error rate (CER \%) evaluated in Parzival datasets in a model obtained after retraining a set of layers in the model previously trained over the IAM database. It has been retrained with a set of 250 lines images}
\LABTAB{Parzival250}
\centering
\begin{tabular}{lrr}
\hline
& \multicolumn{2}{c}{CER (\%)} \\
Trainable Layers &  Validation & Test\\
\hline
BLSTM[12345], FC &  5.2 & 5.4   \\
Conv[5], BLSTM[1,2,3,4,5], FC &  5.1 & 4.8    \\
Conv[4,5]. BLSTM[1,2,3,4,5], FC &  4.4 & 4.5     \\
Conv[3,4,5], BLSTM[1,2,3,4,5], FC &  4.0 & 4.1     \\
Conv[2,3,4,5], BLSTM[1,2,3,4,5], FC &  \textbf{3.6} & \textbf{4.0}   \\
Conv[1,2,3,4,5], BLSTM[1,2,3,4,5], FC &  3.9 & \textbf{4.0}    \\
\hline
\end{tabular}
\end{table}

\begin{table}[!t]
\renewcommand{\arraystretch}{1.3}
\caption{Character error rate (CER \%) evaluated in Parzival datasets in a model obtained after retraining a set of layers in the model previously trained over the IAM database. It has been retrained with a set of 150 lines images}
\LABTAB{Parzival150}
\centering
\begin{tabular}{lrr}
\hline
& \multicolumn{2}{c}{CER (\%)} \\
Trainable Layers &  Validation & Test\\
\hline
BLSTM[12345], FC &  7.4 & 7.4   \\
Conv[5], BLSTM[1,2,3,4,5], FC &  7.2 & 7.0    \\
Conv[4,5]. BLSTM[1,2,3,4,5], FC &  6.8 & 6.5     \\
Conv[3,4,5], BLSTM[1,2,3,4,5], FC &  6.6 & 6.5     \\
Conv[2,3,4,5], BLSTM[1,2,3,4,5], FC &  6.8 & 6.6   \\
Conv[1,2,3,4,5], BLSTM[1,2,3,4,5], FC &  \textbf{6.5} & \textbf{5.8}    \\
\hline
\end{tabular}
\end{table}

\subsection{Data augmentation}
Another additional tool when facing deep learning problems with a small number of labeled data consist in the application of some distortions in the input images in order to augment the database \cite{Yaeger96}. This technique is also applied in \cite{Puigcerver17} over the IAM database, decreasing the CER from $8.2$ \% to $6.4$ \% on the test set and from $5.1$ \% to $4.4$ \% on the validation set. The distortions applied are some affine transformations such as rotation, shearing, translation and scaling, and some morphological distortions such as erosion and dilation. Applying data augmentation to the case of the Washington training set of $150$ images, the CER slightly decrease from $9.4$ to $8.9$ on the test set.

\section{Conclusions}\label{sec:conclusions}
In this paper, TL with a CNN-BLSTM-CTC architecture has been shown to be a promising technique to reduce the number of labeled data when we face an HTR problem over manuscripts belonging to a new domain. Besides the reduced number of labeled data required, this novel procedure  also benefits from a speedup factor since the training is much simpler. In the experiments included, where training over thousands of text lines is transferred to an HTR problem with a few hundred, the proposed TL scheme exhibited a good performance when the whole network is initialized and re-trained. Robust results are obtained if the first or the two first layers of the CNN are kept fixed. Data augmentation can also be used to improve the error. A good performance, with CER in the range $3-9$ \% has been obtained transferring learning from the solution to the HTR of the IAM database to the HTR of the Washington and the Parzival databases, with training data of sizes $150$, $250$ and $350$ and dealing with different resolutions, alphabets and types of images. 

\section*{Acknowledgment}
This  work  was  partially  funded  by  Spanish  government MEC   under   grant   FPU16/04190   and   through   the   project MINECO  TEC2016-78434-C3-2-3-R,  by Comunidad de Madrid under grant IND2017/TIC- 7618, and  by  the  European Union (FEDER). We also gratefully acknowledge the support of  NVIDIA  Corporation  with  the  donation  of  the  Titan  X Pascal GPU used for this research.

%



%



\bibliographystyle{./IEEEtran}
\bibliography{./HTRbib}

\end{document}